%% file: main.tex
\documentclass[10pt,twocolumn,letterpaper]{article}

\usepackage{cvpr}      

\input{preamble}

\definecolor{cvprblue}{rgb}{0.21,0.49,0.74}
\usepackage[pagebackref,breaklinks,colorlinks,citecolor=cvprblue]{hyperref}

\def\paperID{6050} 
\def\confName{CVPR}
\def\confYear{2024}
\title{Hierarchical Patch Diffusion Models for High-Resolution Video Generation}

\author{
Ivan Skorokhodov$^{1,2}$
\quad
Willi Menapace$^{1,3}$
\quad
Aliaksandr Siarohin$^1$
\quad
Sergey Tulyakov$^1$
\\
[-4mm]\and
{Snap Inc.$^1$ \quad KAUST$^2$ \quad University of Trento$^3$}
}

\begin{document}

\maketitle
\input{sec/0_abstract}    
\input{sec/1_intro}
\input{sec/2_related_work}

\input{sec/3_background}
\input{sec/4_method}

\input{sec/5_experiments}

\input{sec/6_conclusion}
{
    \small
    \bibliographystyle{ieeenat_fullname}
    \bibliography{main}
}

\clearpage
\appendix
\setcounter{page}{1}
\maketitlesupplementary

\input{supp/1_limitations}

\input{supp/2_additional_results}
\input{supp/3_impl_details}

\input{supp/4_failed_experiments}
\input{supp/5_societal-impact}

\end{document}

%% file: preamble.tex
\usepackage[dvipsnames]{xcolor}
\usepackage[accsupp]{axessibility} 

\usepackage{amsmath} 
\usepackage{amssymb} 
\usepackage{multirow}
\usepackage{makecell} 
\usepackage{bm} 
\usepackage{stmaryrd} 
\usepackage{dsfont} 
\usepackage{pifont} 
\usepackage{ifthen} 
\usepackage{booktabs} 
\usepackage{soul}
\usepackage{wrapfig} 
\usepackage{colortbl} 

\usepackage{listings}
\usepackage{xcolor}

\definecolor{codegreen}{rgb}{0,0.6,0}
\definecolor{codegray}{rgb}{0.5,0.5,0.5}
\definecolor{codepurple}{rgb}{0.58,0,0.82}
\definecolor{backcolour}{rgb}{0.95,0.95,0.92}

\lstdefinestyle{mystyle}{
    backgroundcolor=\color{backcolour},   
    commentstyle=\color{codegreen},
    keywordstyle=\color{magenta},
    numberstyle=\tiny\color{codegray},
    stringstyle=\color{codepurple},
    basicstyle=\ttfamily\footnotesize,
    breakatwhitespace=false,         
    breaklines=true,                 
    captionpos=b,                    
    keepspaces=true,                 
    numbers=left,                    
    numbersep=5pt,                  
    showspaces=false,                
    showstringspaces=false,
    showtabs=false,                  
    tabsize=2
}
\definecolor{mediumtealblue}{rgb}{0.0, 0.33, 0.71}
\definecolor{darkpastelgreen}{rgb}{0.01, 0.75, 0.24}
\definecolor{azure}{rgb}{0.0, 0.5, 1.0}

\newcommand{\R}{\mathbb{R}}

\newcommand{\x}{\bm{x}}

\newcommand{\ctx}{\text{ctx}}
\newcommand{\act}{\bm{a}}
\newcommand{\coords}{\bm{c}}

\newcommand{\scalef}{s_f}
\newcommand{\scaleh}{s_h}
\newcommand{\scalew}{s_w}
\newcommand{\scales}{\bm{s}}
\newcommand{\offsets}{\bm{\delta}}
\newcommand{\offsetf}{\delta_f}
\newcommand{\offseth}{\delta_h}
\newcommand{\offsetw}{\delta_w}

\newcommand{\fuse}{\texttt{fuse}}
\newcommand{\fvdmini}{FVD@512}
\newcommand{\megapatch}{\bm{P}}
\newcommand{\megasigma}{\bm{\sigma}}

\newcommand{\Normal}{\mathcal{N}}

\newcommand{\apref}[1]{\ref{#1}}
\newcommand{\params}{\bm{\theta}}

\newcommand{\modelfullname}{Hierarchical Patch Diffusion Model}
\newcommand{\modelname}{HPDM}
\newcommand{\modelnameS}{\modelname-S}
\newcommand{\modelnameM}{\modelname-M}
\newcommand{\modelnameL}{\modelname-L}
\newcommand{\modelnameTTV}{\modelname-T2V}
\newcommand{\modelnameTTVK}{\modelname-T2V-1K}

\newcommand{\gridsample}{\texttt{grid\_sample}_\text{3D}}

\newcommand{\noise}{\bm{\varepsilon}}
\newcommand{\patch}{\bm{p}}

\newcommand{\cmark}{\ding{51}}
\newcommand{\xmark}{\ding{55}}

\newcommand{\inlinesection}[1]{\noindent{\textbf{#1}}}

\newcommand{\projecturl}{https://snap-research.github.io/hpdm}
\newcommand{\projecthref}{\href{\projecturl}{\projecturl}}

\newcommand{\expect}[2][]{
\ifthenelse{\equal{#1}{}}{
\mathbb{E}\left[#2\right]
}{
\underset{#1}{\mathbb{E}}\left[#2\right]
}}

\newcommand{\cellbest}{\cellcolor{azure!35}}
\newcommand{\cellsecond}{\cellcolor{azure!10}}

%% file: sec/0_abstract.tex
\begin{abstract}
Diffusion models have demonstrated remarkable performance in image and video synthesis.
However, scaling them to high-resolution inputs is challenging and requires restructuring the diffusion pipeline into multiple independent components, limiting scalability and complicating downstream applications.
In this work, we study patch diffusion models (PDMs) --- a diffusion paradigm which models the distribution of patches, rather than whole inputs, keeping up to ${\approx}$0.7\% of the original pixels.
This makes it very efficient during training and unlocks end-to-end optimization on high-resolution videos.


We improve PDMs in two principled ways.
First, to enforce consistency between patches, we develop \emph{deep context fusion} --- an architectural technique that propagates the context information from low-scale to high-scale patches in a hierarchical manner.
Second, to accelerate training and inference, we propose \emph{adaptive computation}, which allocates more network capacity and computation towards coarse image details.
The resulting model sets a new state-of-the-art FVD score of 66.32 and Inception Score of 87.68 in class-conditional video generation on UCF-101 $256^2$, surpassing recent methods by more than 100\%.
Then, we show that it can be rapidly fine-tuned from a base $36\times 64$ low-resolution generator for high-resolution $64 \times 288 \times 512$ text-to-video synthesis.
To the best of our knowledge, our model is the first diffusion-based architecture which is trained on such high resolutions entirely end-to-end.
Project webpage: \projecthref.
\end{abstract}

%% file: sec/1_intro.tex
\section{Introduction}
\label{sec:intro}

\input{figures/paradigms-comparison}

Recently, diffusion models (DMs) have achieved remarkable performance in image and video synthesis, greatly surpassing previous dominant generative paradigms, such as GANs~\cite{GANs}, VAEs~\cite{VAE} and autoregressive models~\cite{ImageGPT}.
However, scaling them to high-resolution inputs broke their end-to-end nature, since training the full-scale monolithic foundational generator led to infeasible computational demands~\cite{CDM, LDM}.
Splitting the architecture into several stages satisfied the immediate practical needs, but having multiple components in the pipeline makes it harder to tune and complicates downstream tasks like editing or distillation.

For example, LDM~\cite{LDM} trains a diffusion model in the latent space of an autoencoder, which requires an additional extensive hyperparameters search.
The original work has dedicated more than a dozen experiments to it (see Tab. 8 of \cite{LDM}), and the search for its optimal design is still ongoing~\cite{SDXL, DALLE-3, EMU}.
Moreover, retraining an auto-encoder requires retraining the latent generator, resulting in extra computational costs.
Also, having multiple components complicates downstream applications: for example, SnapFusion~\cite{SnapFusion} had to come with two unrelated sets of techniques to distill the generator and the auto-encoder separately.

Cascaded DM (CDM)~\cite{CDM} sequentially trains several diffusion models of increasing resolution, where each next DM is conditioned on the outputs of the previous one.
This framework enjoys a more independent nature of its components, where each generator is trained independently from the rest, but it has more modules in the pipeline (e.g., ImagenVideo~\cite{ImagenVideo} consists of 7 video generators) and more expensive inference.
An end-to-end design is a highly desirable property of a diffusion generator, from the perspectives of both practical importance and conceptual elegance.

The main obstacle to moving a standard high-resolution DM onto end-to-end rails is an increased computational burden.
In the past, patch-wise training proved successful for GAN training for high-resolution image~\cite{ALIS}, video (e.g., \cite{DIGAN, StyleGAN-V}) and 3D (e.g., \cite{GRAF, EpiGRAF}) synthesis, but, however, has not picked up much momentum in the diffusion space.
To our knowledge, PatchDiffusion~\cite{PatchDiffusion} and MaskDIT~\cite{MaskDIT} are the only works that explore it, but none of them considers the required level of input sparsity to scale to high-resolution videos: PatchDiffusion still relies on full-resolution training for 50\% of its optimization (so it is not purely patch-wise), while MaskDIT preserves ${\approx}$50\% of the original input.
In our work, we explore patch diffusion models while keeping just \emph{up to 0.7\%} of the original pixels.
The comparison of patch-wise training and conventional paradigms is depicted in Fig.~\ref{fig:paradigms-comparison}, and in Table~\ref{tab:efficiency}, we show that it can achieve $\times 5$ larger throughput and is trainable on high-resolution videos.
We focus on video synthesis since, for videos, the computational burden of high resolutions is considerably more pronounced than for images: there now exist end-to-end image diffusion models that are able to train even on $1024^2$ resolution (e.g., \cite{simple-diffusion, MDM, VDM++,HourglassDiffusion}).


\input{tables/efficiency}

For our patch-wise training, we consider a hierarchy of patches instead of treating them independently~\cite{PatchDiffusion}, which means that the synthesis of high-resolution patches is conditioned on the previuosly generated low-resolution ones.
It is a similar idea to cascaded DMs~\cite{CDM} and helps to improve the consistency between patches and simplifies noise scheduling for high resolutions~\cite{RelayDiffusion,simple-diffusion,ImportanceOfNoiseScheduling}. 
To improve both the qualitative performance and computational efficiency of patch diffusion, we develop two principled techniques: deep context fusion and adaptive computation.

Deep context fusion considers conditions the generation of higher-resolution patches on subsampled, positionally aligned features from the lower levels of the pyramid.
It serves as an elegant way to incorporate global context information into synthesis of higher-frequency textural details and to facilitate knowledge sharing between the stages.
Adaptive computation restructures the model architecture in such a way that only a subset of layers operate on high-resolution patches, while more difficult low-resolution ones go through the whole pipeline.

We apply the designed techniques to the recent attention-based RIN generator~\cite{RIN}, and benchmark our approach on two video generation datasets: UCF-101~\cite{UCF101_dataset} in the $64 \times 256^2$ resolution, and our internal dataset of text/video pairs for $64\times288\times512$ (and $16\times576\times1024$) text-to-video generation.
Our model achieves state-of-the-art performance on UCF-101 and demonstrates strong scalability performance for large-scale text-to-video synthesis.

%% file: figures/paradigms-comparison.tex
\begin{figure}[t!]
\begin{center}
\includegraphics[width=\linewidth]{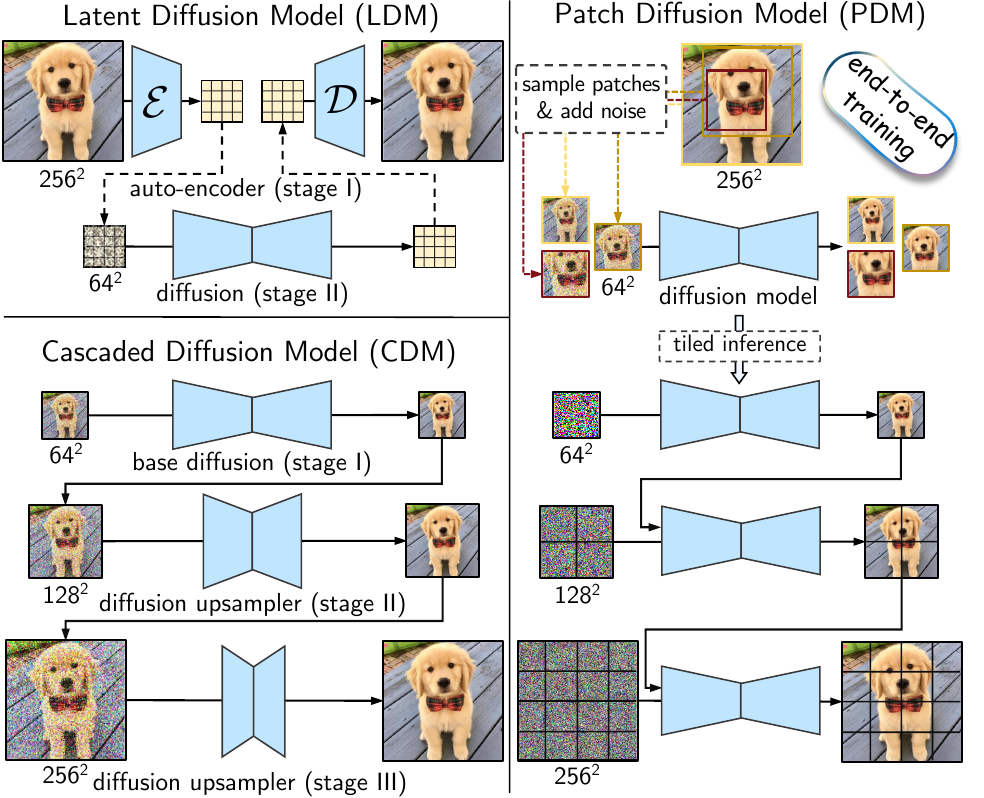}
\end{center}
\caption{Comparing existing diffusion paradigms: Latent Diffusion Model (LDM)~\cite{LDM,LSGM} (upper left), Cascaded Diffusion Model (CDM)~\cite{CDM} (bottom left), and Patch Diffusion Model (this work) during training (upper right) and inference (bottom right). In our work, we develop \emph{hierarchical} patch diffusion, which never operates on full-resolution inputs, but instead optimizes the lower stages of the hierarchy to produce spatially aligned context information for the later pyramid levels to enforce global consistency between patches.}
\label{fig:paradigms-comparison}
\end{figure}

%% file: tables/efficiency.tex
\begin{table}[t]
\caption{Efficiency comparison between patch-wise and full-resolution diffusion in the RIN~\cite{RIN} framework (which scales more gracefully with the input size than UNets~\cite{U-net, ADM}). Memory consumption is measured in GB for the batch size of 1; speed as videos/sec for a maxed-out batch size on NVidia A100 80GB.}
\label{tab:efficiency}
\centering
\resizebox{\linewidth}{!}{
\begin{tabular}{lcccc}
\toprule
\multirow{2}{*}{Method} & \multicolumn{2}{c}{$64 \times 256^2$} & \multicolumn{2}{c}{$64 \times 512^2$}  \\
& Mem $\downarrow$ & Speed $\uparrow$ & Mem $\downarrow$ & Speed $\uparrow$ \\
\midrule
Full-resolution DM & 65.3 & 1.24 & OOM & OOM \\
HPDM ($32\times128^2$ patch size) & 29.0 & 2.64 & 41.3 & 1.55 \\
~+ adaptive computation & 23.4 & 3.58 & 29.9 & 2.49 \\
HPDM ($16\times64^2$ patch size) & 18.1 & 4.25 & 22.1 & 2.78 \\
~+ adaptive computation & 14.2 & 6.71 & 16.3 & 4.96 \\
\bottomrule
\end{tabular}
}
\end{table}

%% file: sec/2_related_work.tex
\section{Related work}
\label{sec:related-work}

\inlinesection{High-level diffusion paradigms}.
To the best of our knowledge, one can identify two main conceptual paradigms on how to structure a high-resolution diffusion-based generator: latent diffusion models (LDM)~\cite{LDM} and cascaded diffusion models (CDM)~\cite{CDM}.
For CDMs, it was shown that the cascade can be trained jointly~\cite{f-DM}, but scaling for high resolutions or videos still requires progressive training from low-resolution models to obtain competitive results~\cite{MDM}.

\inlinesection{Video diffusion models}.
The rise of diffusion models as foundational image generators~\cite{ADM,DALLE-2} motivated the community to explore them for video synthesis as well~\cite{VideoDiffusionModels}.
VDM~\cite{VideoDiffusionModels} is one of the first works to demonstrate their scalability for conditional and unconditional video generation using the cascaded diffusion approach~\cite{CDM}.
ImagenVideo~\cite{ImagenVideo} further pushes their results, achieving photorealistic quality.
VIDM~\cite{VIDM} designs a separate module to implicitly model motion.
PVDM~\cite{PVDM} trains a diffusion model in a spatially decomposed latent space.
Make-A-Video~\cite{Make-A-Video} uses a vast unsupervised video collection in training a text-to-video generator by fine-tuning a text-to-image generator.
PYoCo~\cite{PYoCo} and VideoFusion~\cite{VideoFusion} design specialized noise structures for video generation.
Numerous works explore training of a foundational video generator on limited resources by fine-tuning a publicly available StableDiffusion~\cite{LDM, SDXL} model for video synthesis (e.g., ~\cite{VideoFusion, VideoFactory, AnimateDiff, ZeroScope, VideoLDM}).
Another important line of research is the adaptation of the foundational image or video generators for downstream tasks, such as video editing (e.g. \cite{Text2VideoZero,TokenFlow,Tune-a-video,Gen-1,Rerender-A-Video}) or 4D generation~\cite{Make-It-4D}.
None of these models is end-to-end and all follow cascaded~\cite{CDM} or latent~\cite{LDM} diffusion paradigms.



\inlinesection{Patch Diffusion Models}.
Patch-wise generation has a long history in GANs~\cite{GANs} and has enjoyed applications in image~\cite{COCO-GAN}, video~\cite{StyleGAN-V} and 3D synthesis~\cite{GRAF}.
In the context of diffusion models, there are several works that explore patch-wise inference to extend foundational text-to-image generators to higher resolutions than what they had been trained on (e.g., \cite{MultiDiffusion, AnySizeDiffusion, CSD}).
Also, a regular video diffusion model can be inferred in an autoregressive manner at the test time because it can be easily conditioned on its previous generations via classifier guidance or noise initialization~\cite{VideoDiffusionModels}, and this kind of synthesis can also be seen as a patch-wise generation.
Later stages of CDMs can also operate in a patch-wise fashion~\cite{DALLE-2}, even though they have not been explicitly trained for this.
These works have relevance to ours, since they design patch-wise sampling strategies with better global consistency in the resulting samples and thus could be employed for our generator as well.

The primary focus of our work is patch-wise training of diffusion models, which has been explored in several prior works.
Several works (e.g., \cite{SinDiffusion, SinFusion, SinDDM}) train a diffusion model on a single image to produce its variations~\cite{SinGAN, DropTheGAN}.
The closest work to ours is PatchDiffusion~\cite{PatchDiffusion}, which explores direct patch-wise diffusion training.
However, to learn the consistent global image structure, their developed model operates on full-size inputs in 50\% of the optimization steps, which is computationally infeasible for high-resolution videos.
Our generator design, in contrast, \emph{never} operates on full-resolution videos and instead relies on context fusion to enforce the consistency between the patches.

Apart from expensive training, diffusion models also suffer from slow inference~\cite{ADM}, and some works explored alternative denoising paradigms (e.g., \cite{MAGVIT, MAGVITv2}) to mitigate this, which is a close but orthogonal line of research.





\input{figures/architecture}

%% file: figures/architecture.tex
\begin{figure}
\centering
\includegraphics[width=0.8\linewidth]{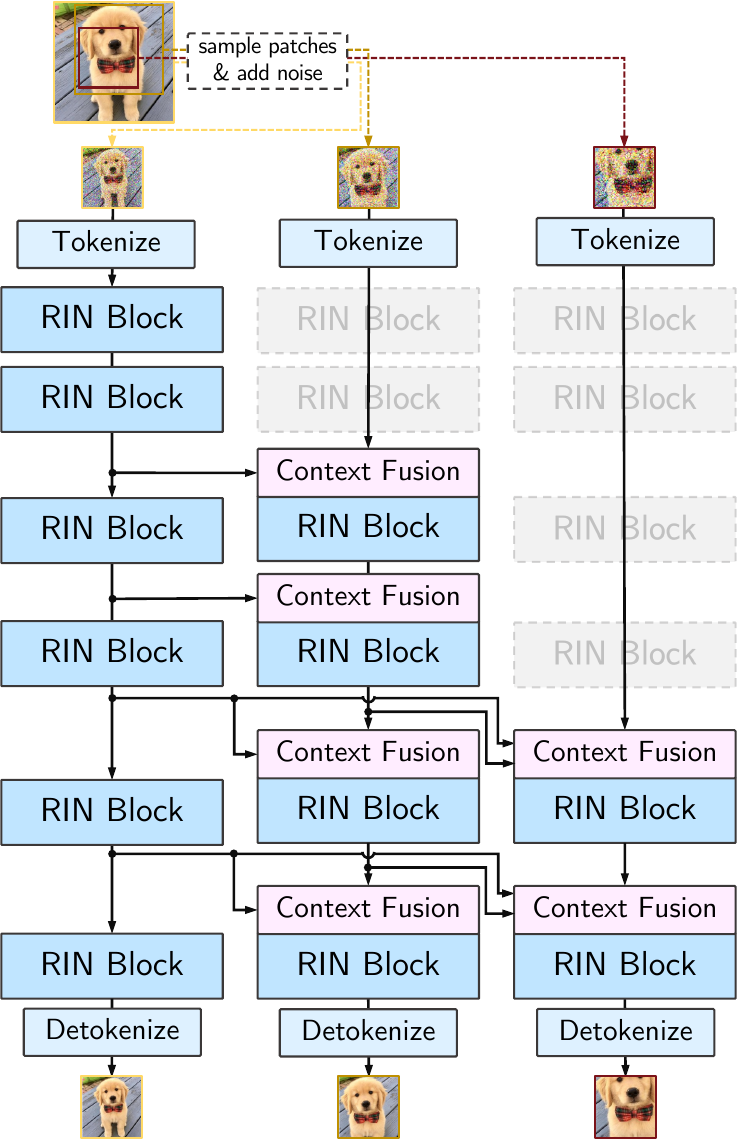}
\caption{Architecture overview of \modelfullname\ (\modelname) for a 3-level pyramid. The model is trained to denoise all the patches jointly. During training, we use only a single patch from each pyramid level and restrict information propagation in the coarse-to-fine manner. This allows one to synthesize the whole image (or video) at a given resolution patch-by-patch using tiled inference (see Figure~\ref{fig:paradigms-comparison}).}
\label{fig:architecture}
\end{figure}

%% file: sec/3_background.tex
\section{Background}
\label{sec:background}

\subsection{Diffusion Models}
Given a dataset $\bm{X} = \{\bm x^{(n)}\}_{n=1}^N$, consisting of $N$ samples $\bm{x}^{(n)} \in \R^d$ (most commonly images or videos), we seek to recover the underlying data-generating distribution $\bm{x}^{(n)} \sim p(\bm x)$.
We follow the general design of time-continuous diffusion models~\cite{EDM}, for which a neural network $D_{\params}(\tilde{\x}; \sigma)$ is trained to predict ground-truth dataset samples $\x$ from their noised versions $\tilde{\x} = \x + \noise, \noise \sim \Normal(\bm 0, \sigma\bm{I})$:
\begin{align}
\expect[\x, \noise, \sigma]{\|D_{\params}(\tilde{\x}; \sigma) - \x\|_2^2} \rightarrow \min_{\params}
\end{align}
In the above formula, $p(\sigma)$ controls the corruption intensity and its distribution parameters are treated as hyperparameters~\cite{ImportanceOfNoiseScheduling,EDM}.
The denoising network can serve as a score estimator~\cite{EDM}:
\begin{align}
\bm{s}_{\params}(\bm x, \sigma) \triangleq \nabla_{\x} \log p(\x) \approx \frac{1}{\sigma^2}(D_{\params}(\x; \sigma) - \x).
\end{align}
For large enough $\sigma$, the corrupted sample $\tilde{\x}$ is indistinguishable from pure Gaussian noise, and this allows to employ the score predictor for sampling at test-time using Langevin dynamics~\cite{NCSN} (with $\sigma \rightarrow 0$ and $T \rightarrow \infty$):
\begin{align}
\tilde{\x}_t = \tilde{\x}_{t-1} + \frac{\sigma}{2} \bm{s}_{\params}(\tilde{\x}, \sigma) + \noise_t.
\end{align}






\subsection{Recurrent Interface Networks}
For our base architecture, we chose to follow Recurrent Interface Networks (RINs)~\cite{RIN} for their simplicity and expressivity.
A typical RIN network has a uniform structure and consists of a ViT-like~\cite{ViT} linear image tokenizer, followed by a sequence of identical attention-only blocks and a linear detokenizer to transform the image tokens back to the RGB pixel values.
RIN blocks do not employ an expensive self-attention mechanism~\cite{Transformer} and instead rely on linear cross-attention layers with a set of learnable latent tokens.
This allows to scale gracefully with input resolution without sacrificing communication between far-away input locations.
We refer the reader to the original work~\cite{RIN} for additional details and provide the illustration for our RIN block in Figure~\apref{fig:architecture-full} in Appendix~\apref{sec:impl-details}.

%% file: sec/4_method.tex
\section{Method}
\label{sec:method}

Our high-level patch diffusion design is different from PatchDiffusion~\cite{PatchDiffusion} in that our model never operates on full-resolution inputs.
Instead, we consider a hierarchical cascade-like structure consisting of $L$ stages and patch scales $s_\ell$ decrease exponentially: $s_\ell = 1/2^{\ell}$ for $\ell \in \{0, 1, ..., L\}$.
Patches are always of the same resolution $\bm{r} = (r_f, r_h, r_w)$, which leads to substantial memory and computational savings compared to full-resolution training.
During training, we randomly sample a video from the dataset and extract a hierarchy of patch coordinates $\coords_0, ..., \coords_L$ in such a way that the $\ell$-th patch is always located inside the previous $\ell'<\ell$ patches so that they provide the necessary context information.
Hierarchical patch diffusion is trained to jointly denoise a combination of these patches, denoted as $\megapatch = (\patch\ell){\ell=0}^L$, and their corresponding noise levels $\megasigma = (\sigma\ell)_{\ell=0}^L$:
\begin{align}
\expect[\patch, \noise, \sigma]{\|D_{\params}(\tilde{\megapatch}_0; \megasigma) - \megapatch\|_2^2} \rightarrow \min_{\params},
\end{align}
where each patch is corrupted independently: $\tilde{\megapatch} = (\tilde{\patch}_\ell + \noise_\ell)_{\ell=0}^L, \noise_\ell \sim \Normal(0, \sigma_\ell, I)$
Restricting the information flow in the coarse-to-fine manner (see \cref{fig:architecture}) allows to do inference at test-time in the cascaded diffusion fashion~\cite{CDM}.

Below, we elaborate on three fundamental components of our method that allow a patch-wise paradigm to achieve state-of-the-art results in video generation: deep context fusion, adaptive computation and overlapped sampling.





\subsection{Patch Diffusion}
The training objective of patch diffusion is similar to the regular diffusion design, but instead of full-size videos (or images) $\x \in \R^{R_f \times R_h \times R_w}$, it uses randomly subsampled patches $\patch \in \R^{r_f \times r_h \times r_w}$ and trains the patch-wise model $D_{\params}(\tilde{\patch}; \sigma)$ to denoise them:
\begin{align}
\expect[\patch, \noise, \sigma]{\|D_{\params}(\tilde{\patch}; \sigma) - \patch\|_2^2} \rightarrow \min_{\params}.
\end{align}
Following \cite{EpiGRAF}, the patch extraction procedure extracts pixels using random scales $\scales = (\scalef, \scaleh, \scalew)$, $s_* \in [r_*/R_*, 1]$, and offsets $\offsets = (\offsetf, \offseth, \offsetw), \delta_* \in [0, 1 - s]$:
\begin{align}
\patch = \texttt{downsample}(\texttt{crop}(\x; \bm{\delta}); \bm{r}),
\end{align}
where the \texttt{crop} function slices the input signal given the pixel offsets $\bm\delta$, and $\texttt{downsample}$ resizes it to the specified resolution $r_f \times r_h \times r_w$.

Since we consider a hierarchical structure, during training, we use fixed scales for each $\ell$-th level $s^{(\ell)}_* = r_*^\ell / R_*$, but randomly sample offsets $\delta_*^{(\ell)} \sim U[0, 1 - s_*^{(\ell)}]$.
For a level $\ell > 1$, we sample its corresponding offset $\delta_*^{(\ell)}$ in each $*$-th dimension in such a way, that the resulting patch is always located inside the patch from the previous pyramid level, as visualized in \cref{fig:architecture}.
For brevity, we will omit the level superscript in the subsequent exposition for patch parameters.


Setting patch resolutions $r_f, r_h, r_w$ lower than original ones $R_f, R_h, R_w$ leads to drastic improvements in computational efficiency, but worsens the global consistency of the generated samples.
In \cite{PatchDiffusion}, the authors use variable-resolution training, including 50\% of optimization steps performed on full-size inputs to improve the consistency.
The downside of such a strategy is that it undermines computational efficiency: for a large enough video, the model cannot fit into GPU memory even for a batch size of 1.
Instead, in our work, we demonstrate that consistent generation can be achieved with \textit{deep context fusion}: conditioning higher resolution generation on the activations from previously generated stages.

\subsection{Deep Context Fusion}
\label{sec:method:context-fusion}
\input{figures/context-fusion}
The main struggle of patch-wise models is preserving the consistency between the patches, since each patch is modeled independently from the rest, conditioned on the previous pyramid stage.
Cascaded DMs~\cite{CDM} provide the conditioning to later stages by simply concatenating an upsampled low-resolution video channel-wise~\cite{CDM} to the current latent.
While it can provide the global context information when the model operates on a full-resolution input, for patch-wise models, this leads to drastic context cut-outs, which, as we demonstrate in our experiments, severely worsens the performance.
Also, it limits the knowledge sharing between lower and higher stages of the cascade.
To address this issue, we introduce \emph{deep context fusion (DCF)}, a context fusion strategy that conditions the higher stages of the pyramid on spatially aligned, globally pooled features from the previous stages.

For this, before each RIN block of our model, we pool the global context information from previous stages into its inputs.
For this, we use the patch coordinates to grid-sample the activations with trilinear interpolation from all previous pyramid stages, average them, and concatenate to the current-stage features.

More precisely, for a given patch $b$-th block inputs $\act_{\ell}^{b-1} \in \R^{d\times r'_f \times r'_h \times r'_w}$ with coordinates $\coords_\ell = (s, \offsetf, \offseth, \offsetw) \in \R^4$ at the $\ell$-th pyramid level; $\ell-1$ context patches' activations  $(\act_k^{b-1})_{k=1}^{\ell-1}$ with respective coordinates $(\coords_k)_{k=1}^{\ell-1}$, we compute the context $\ctx_\ell \in \R^{d\times r'_f \times r'_h \times r'_w}$ as:
\begin{align}
\ctx_\ell^b = \frac{1}{\ell-1}\sum_{k=1}^{\ell-1} \gridsample[\act_{\ell-1}^{b-1}, \hat{\coords}_{\ell}],
\end{align}
where $\gridsample$ is a function that extracts the features with trilinear interpolation via the coordinates queries, $\hat{\coords}_{\ell}$ are the recomputed patch coordinates (for $k < \ell$) calculated as:
\begin{align}
\hat{\coords}_{\ell}(\coords_\ell, \coords_k) = [s_\ell / s_k; (\offsets_\ell - \offsets_k) / s_k].
\end{align}
We fuse this context information via simple channel-wise concatenation together with the coordinates information $\coords_\ell$ which we found to be slightly improving the consistency:
\begin{align}
\label{eq:context-fusion}
\fuse[\act^{b-1}_{\ell}, \coords_\ell; (\act_k^{b-1}, \coords_k)_{k=1}^{\ell-1}] = \texttt{concat}[\act_{\ell}, \ctx_\ell, \coords_\ell].
\end{align}
Deep context fusion is illustrated in \cref{fig:context-fusion}.

To keep the dimensionalities the same across the network, we then project the resulted tensor $\fuse[\cdot] \in \R^{(2d + 3) \times r'_f \times r'_h \times r'_w}$ with a learnable linear transformation.
We considered other aggregation strategies, like concatenating all the levels's features or averaging, but the former one blows up the dimensionalities, making the training expensive, while the latter one was leading to poor performance in our preliminary experiments.

An additional advantage of DCF compared to shallow context fusion of regular cascaded DMs is that the gradient can flow from the small-scale patch denoising loss to the lower levels of the hierarchy, pushing the earlier cascade stages to learn such features that are more useful to the later ones.
We found that this is indeed helpful in practice and improves the overall performance additionally by ${\approx}5\%$.


\input{figures/ucf-comparison}

\subsection{Adaptive Computation}
\label{sec:method:adaptive-computation}

Naturally, generating high-resolution details is considered to be easier than synthesizing the low-resolution structure~\cite{SWAGAN}.
In this way, allocating the same amount of network capacity on high-resolution patches can be excessive, that is why we propose to use only some of the computational blocks when running the last stages of the pyramid.
We name this strategy \emph{adaptive computation}\footnote{Our notion of adaptive computation is different from the original RIN's one, where it is used to describe the model's ability to distribute its computational capacity differently between different parts of an input~\cite{RIN}.} and demonstrate that it improves our model's efficiency by ${\approx}60\%$ without compromising the performance (see \cref{tab:ucf-ablations}).
The uniform RIN's structure~\cite{RIN} (i.e., all the blocks are identical and have the same input/output resolutions) allows us to implement this easily: one simply skips some of the earlier blocks when processing the high-resolution activations.
The high-level pseudo-code is provided in \cref{alg:adaptive-computation}.

\input{algorithms/adaptive-computation}

Adaptive computation involves two design choices: 1) whether to skip earlier or later blocks in the networks for higher resolutions, and 2) how to distribute the computation assignments among the blocks per each pyramid stage.
We chose to allocate the later blocks to perform full computation to make the low-level context information go through more processing before being propagated to the higher stages.
For the block allocations, we observed that simply increasing the computation assignments linearly with the block index worked well in practice.

\subsection{Tiled Inference}
\label{sec:method:sampling}
Sampling from HPDM is different from regular diffusion sampling, since it is patch-wise and we never operate on full-resolution inputs.
During inference, we generate pyramid levels one-by-one, starting from $r_t \times r_h \times r_w$ video (corresponding to a patch of scale $s = 1$), then using to generate the video of resolution $2r_t \times 2r_h \times 2r_w$ (corresponding to patch scale $s = 1/2$), and so on until we produce the final video of full resolution $R_f \times R_h \times R_w$.
We visualize this hierarchical tiled inference process in \cref{fig:paradigms-comparison} (bottom right).

Each next stage of the pyramid uses the generated video from the previous stage through the deep context fusion technique described in \cref{sec:method:context-fusion}.
DCF provides strong global context conditioning, but it is sometimes not enough to enforce local consistency between two neighboring patches.
To mitigate this, we employ the MultiDiffusion~\cite{MultiDiffusion} strategy and simply average-overlap the score predictions $\bm{s}_{\params}(\hat{\patch}, \sigma)$ during the denoising process.
More concretely, to generate a complete video $\x \in \R^{R_f \times R_h \times R_w}$, we first generate $(2R_f - 1) \times (2R_h - 1) \times (2R_w - 1)$ patches with 50\% of the coordinates overlapping between two neighboring patches.
Then, we run the reverse diffusion process for each patch and average the overlapping regions of the corresponding score predictions.
The importance of overlapped inference is illustrated in \cref{fig:overlapped-inference} and \cref{tab:ucf-inference-ablations}.

\subsection{Miscellaneous techniques}
\label{sec:method:misc}

The core ideas that enable our work have been described above, but from the implementation and engineering standpoints, there are several other techniques that played an important role in bolstering the performance and would be of interest to a practitioner aiming to reproduce our results.
Additional details and failed experiments can be found in Appendix~\apref{supp:sec:limitations} and \apref{supp:sec:failed-experiments}, respectively.

\inlinesection{Integer patch coordinates}.
We noticed that sampling a patch on the $L$-th cascade level at integer coordinates allows to prevent blurry artifacts in generated videos: they appear due to oversmoothness effects of trilinear interpolation.

\inlinesection{Noise Schedule}
Each stage of the pyramid operates on different frequency signals, and higher levels of the pyramid have stronger correlations between patch pixels.
Inspired by \cite{ImportanceOfNoiseScheduling}, we found it helpful to use exponentially smaller input noise scaling with each increase in pyramid level.

\inlinesection{Cached inference}
During inference, we do not need to recompute all the activations for the previous pyramid stages, which makes it possible to cache them, which works even more gracefully.
Caching block features allowed to speed up the inference by ${\approx}$40\%.
However, for the large model, caching needs to be implemented with CPU offloading to prevent GPU out-of-memory errors.

\subsection{Implementation details}
\label{sec:method:details}

We use RINs~\cite{RIN} instead of U-Nets~\cite{U-net, ADM} as the backbone since its uniform structure is conceptually simpler and aligns well with adaptive computation.
We use $\bm{v}$-prediction parametrization~\cite{v-prediction} with extra input scaling~\cite{ImportanceOfNoiseScheduling}.
Following RINs, we train our model with the LAMB optimizer~\cite{LAMB}, with the cosine learning rate schedule and the maximum LR of 0.005.
Our model has 6 RIN blocks, and we distribute the load for adaptive computation as [1, 1, 2, 2, 3, 4]: e.g., the 1-st and 2-nd blocks only compute the first pyramid level, the 3-rd and 4-rd ones --- first two levels of the pyramid, and so on.
Not using adaptive computation is equivalent to having a load of [4, 4, 4, 4, 4, 4], which is almost twice as expensive.
We use 768 latent tokens of 1024/3072 dimensionality with $1\times 4 \times 4$ pixel tokenization for class-conditional/text-conditional experiments, respectively.
To encode the textual information, we rely on T5 language model~\cite{T5} and use its T5-11B variant.
Further implementation details can be found in Appx~\apref{sec:impl-details}.

%% file: figures/context-fusion.tex
\begin{figure}[t]
\centering
\includegraphics[width=0.8\linewidth]{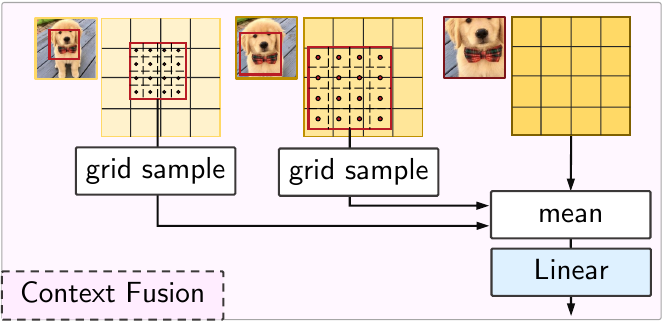}
\caption{\textbf{Deep Context Fusion}. At each pyramid level, we grid-sample the features of a lower-resolution patch and concatenate them to the activations tensor of the current level. In this way, the information propagates in the coarse-to-fine manner and provides richer context than pixel-space concatenation of cascaded DMs (see \cref{tab:ucf-ablations}).}
\label{fig:context-fusion}
\end{figure}

%% file: figures/ucf-comparison.tex
\begin{figure*}
\centering
\centering
\begin{subfigure}[b]{0.49\textwidth}
    \centering
    \includegraphics[width=\textwidth]{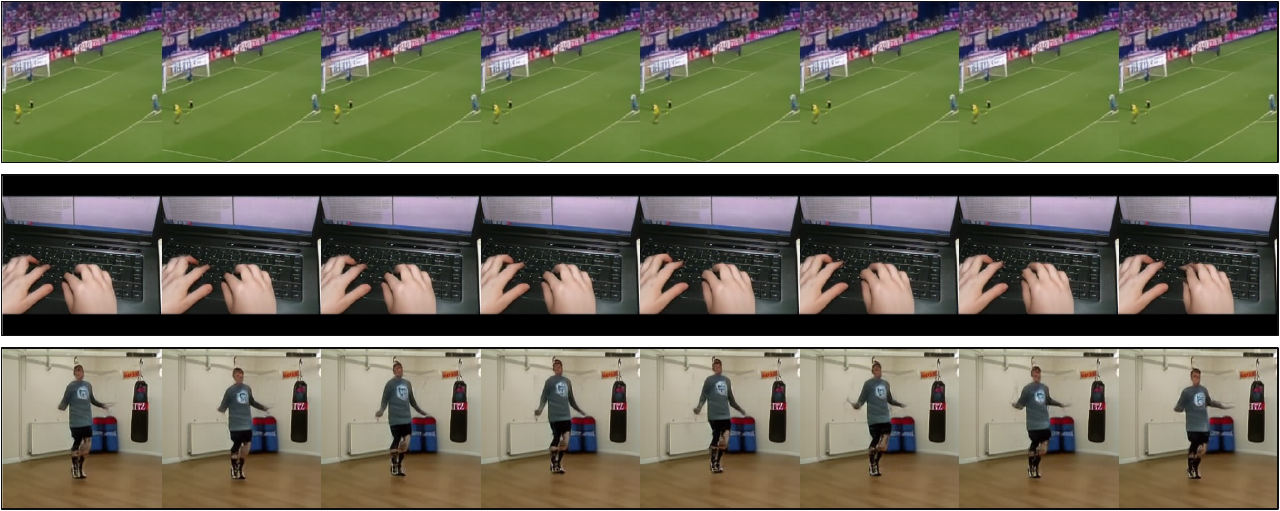}
\end{subfigure}
\hfill
\begin{subfigure}[b]{0.49\textwidth}
    \centering
    \includegraphics[width=\textwidth]{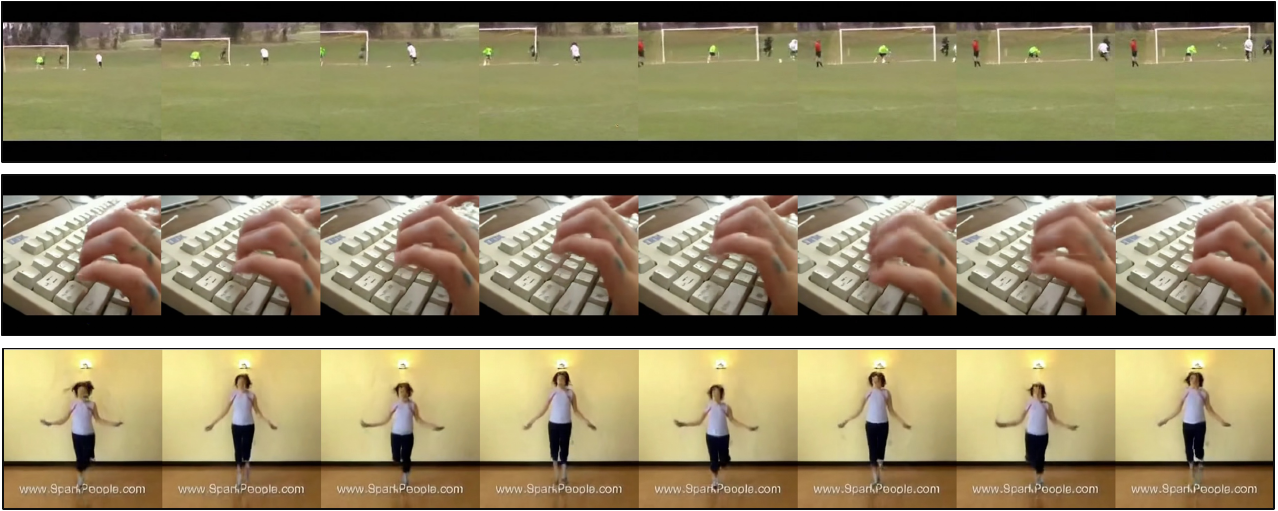}
\end{subfigure}
\caption{Provided samples from PVDM~\cite{PVDM} (left) and random samples from \modelnameL\ (right) for the same classes on UCF $256^2$. More samples are provided in Appendix~\apref{supp:sec:additional-results}.}
\label{fig:ucf-comparison}
\end{figure*}

%% file: algorithms/adaptive-computation.tex
\begin{lstlisting}[language=Python, caption={Pseudo-code for adaptive computation (\cref{sec:method:adaptive-computation})},label=alg:adaptive-computation]
def adaptive_computation(
    blocks: List[RINBlock],
    x: Tensor,
    num_levels_per_block: List[int]
  ) -> Tensor:
  # `x` has the shape: [B, L, D, F, H, W]
  for blk_idx, blk in enumerate(blocks):
    nlvl: int = num_levels_per_block[blk_idx]
    x[:, :nlvl] = blk(x[:, :nlvl])
\end{lstlisting}

%% file: sec/5_experiments.tex
\section{Experiments}
\label{sec:experiments}

\input{figures/large-model-results}


\inlinesection{Datasets}.
In our work, we consider two datasets: 1) UCF101~\cite{UCF101_dataset} (for exploration and ablations) and 2) our internal video dataset to train a large-scale text-to-video model.
UCF101 is a popular academic benchmark for unconditional and class-conditional video generation consisting of videos of the $240 \times 320$ resolution with 25 FPS and has an average video length of ${\approx}7$ seconds.
Our internal dataset consists of ${\approx}$25M high-quality text/video pairs in the style of stock footage with manual human annotations and ${\approx}70M$ of low-quality in-the-wild videos with automatically generated captions.
Additionally, for text-to-video experiments, we used an internal dataset of ${\approx}$150M high-quality text/image pairs for extra supervision~\cite{Make-A-Video}.

\inlinesection{Evaluation}.
Following prior work~\cite{ImagenVideo, StyleGAN-V, CogVideo, VideoDiffusionModels}, we evaluate the model with two main video quality metrics: Frechet Video Distance (FVD)~\cite{FVD}, and Inception Score (IS)~\cite{TGANv2}.
For FVD and IS, we report their values based on 10,000 generated videos.
But for ablations, we use \fvdmini\ instead for efficiency purposes: an FVD variant computed on just 512 generated videos.
We noticed that it correlates well with the traditional FVD, but with just a fixed offset.
Apart from that, we report the training throughput for various designs of our network and also provide the samples from our model for qualitative assessment.


\subsection{Video generation on UCF-101}

We train \modelname\ in three variants: \modelnameS\, \modelnameM, and \modelnameL, which differ in the amount of parameters, batch size and training iterations used.
The hyperparameters for them are provided in \cref{tab:hyperparameters}.
For ablations, we train all the models for 50K steps with the batch size of 512.
UCF models are trained for the final video resolution of $64\times256^2$ with the pyramid $16\times64^2 \rightarrow 32\times128^2 \rightarrow 64\times256^2$.

\inlinesection{Main results}.
Our patch-wise model is trained on UCF-101~\cite{UCF101_dataset} for $64\times 256^2$ generation entirely end-to-end with the hierarchical patch sampling procedure described in \cref{sec:method}.
In \cref{tab:ucf-zero-shot}, we compare these results with recent state-of-the-art methods: MoCoGAN-HD~\cite{MoCoGAN-HD}, StyleGAN-V~\cite{StyleGAN-V}, TATS~\cite{TATS}, VIDM~\cite{VIDM}, DIGAN~\cite{DIGAN}, PVDM~\cite{PVDM}.
While our model is trained to synthesize 64 frames, we report quantitative results for 16 generated frames, since it is a much more popular benchmark in the literature (for this, we simply subsample 16  frames out of the generated 64).
Our model substantially outperforms all previously reported results for this benchmark (i.e., for the $16 \times 256^2$ resolution and without pretraining) by a striking margin of more than $100\%$.
To our knowledge, these are the best reported FVD and IS scores for the $16 \times 256^2$ resolution on UCF.
Make-A-Video~\cite{Make-A-Video} reports FVD of 81.25 and IS of 82.55 when fine-tuned from a large-scale text-to-video generator.

\inlinesection{Ablations}.
We consider two lines of ablations: ablating core architectural decisions and benchmakring various inference strategies, since the latter also crucially influences the final performance.
For the training components, we first analyze the influence of deep context fusion.
For this, we launch an experiment with ``shallow context fusion'', where we concatenate only the RGB pixels (non-averaged, only from the patch of the previous pyramid level) as the context information.
As one can see from the results in \cref{tab:ucf-ablations} (first row), this strategy produces considerably worse results (though the training becomes ${\approx}10\%$ faster).

The next ablation is whether the low-level pyramid stages indeed learn such features that are more useful for later pyramid stages, when they are directly supervised with the denoising loss of small-scale patches through the context aggregation procedure.
For this ablation, we detach the context variable $\ctx$ from the autograd graph.
The results are presented in \cref{tab:ucf-ablations} (second row).
One can observe that the performance can be better for earlier pyramid stages, but the late stage suffers: this demonstrates that the lowest stage indeed learns to encode the global context in a way that is more accessible for later levels of the cascade, but by sacrificing a part of its capacity due to this.

One of the key techniques we used in our model is adaptive computation, and in \cref{tab:ucf-ablations} (third row), we demonstrate how the model performs without it.
While it allows to obtain slightly better results, it decreases the training speed by almost twice.
The cost of the later pyramid stages becomes even more critical during inference time, when sampling high-resolution videos.

Finally, we verify the existing observation of the community that positional encoding in patch-wise models help in producing more spatially consistent samples~\cite{COCO-GAN, ALIS}.
This can be seen from the worse \fvdmini\ scores in \cref{tab:ucf-ablations} (4th row) when no coordinates information is input to the model in context fusion (\cref{eq:context-fusion}).

\input{tables/ucf-main}
\input{tables/ucf-ablations}
\input{tables/ucf-inference-ablations}
\input{figures/overlapped-inference}

\subsection{Text-to-video generation}
\inlinesection{Training setup}.
To explore the scalability of the patch-wise paradigm, we launched a large-scale experiment for \modelname\ with ${\approx}$4B parameters on a text/video dataset consisting of ${\approx}$95M samples.
Since training a foundational model incurs extreme financial costs, we instead found it financially less risky to fine-tune it from a low-resolution generator.
For this, we used the base SnapVideo~\cite{SnapVideo} model, which operates on $36 \times 64$ resolution videos.
Our patch-wise variant, \modelnameTTV, was trained for the final output resolution of $64 \times288 \times 512$ with the pyramid $8 \times 36 \times 64 \rightarrow 16 \times 72 \times 128 \rightarrow 32 \times 144 \times 256 \rightarrow 64 \times 288 \times 512$ (4 pyramid levels in total).
This 4-level pyramid structure results in just $4 \cdot (1/8)^3~{\approx}~0.7$\% of the original video pixels seen in each optimization step.
The base $36 \times 64$ generator was trained for 500,000 iterations, and we fine-tuned \modelnameTTV\ for 15,000 more steps (3\% of the base generator training steps) with a batch size of 4096.
We also fine-tune another model, \modelnameTTVK, a $16\times576\times1024$ text-to-video generator with a patch resolution of $16\times 72 \times 128$.
It is initialized from the base $36\times64$ SnapVideo diffusion model, but fine-tuned for 100,000 iterations.
Longer fine-tuning was required for it since its input resolution was chosen to be larger than that of the base generator to make it have 4 levels in the pyramid instead of 5.
Apart from videos, following prior works (e.g., \cite{VideoDiffusionModels, Make-A-Video}), we utilize joint image/video training.
For image training with RINs, following SnapVideo~\cite{SnapVideo}, we simply repeat the image along the time axis to convert it into a still video.

\inlinesection{Results}.
We test the results quantitatively by reporting zero-shot performance on UCF-101~\cite{UCF101_dataset} in terms of FVD and IS in \cref{tab:ucf-zero-shot}, and also qualitatively by providing visual comparisons with existing foundational generators in \cref{fig:large-model-results}.
Although trained for just 15,000 steps, \modelnameTTV\ yields promising results and has a comparable generation quality to modern foundational text-to-video models (ImageVideo~\cite{ImagenVideo}, Make-A-Video~\cite{Make-A-Video}, and PYoCo~\cite{PYoCo}) on some text prompts (see \cref{fig:large-model-results}).

We provide more qualitative results on the project webpage: \projecthref.

\input{tables/ucf-zero-shot}

%% file: figures/large-model-results.tex
\begin{figure*}
\centering
\includegraphics[width=\linewidth]{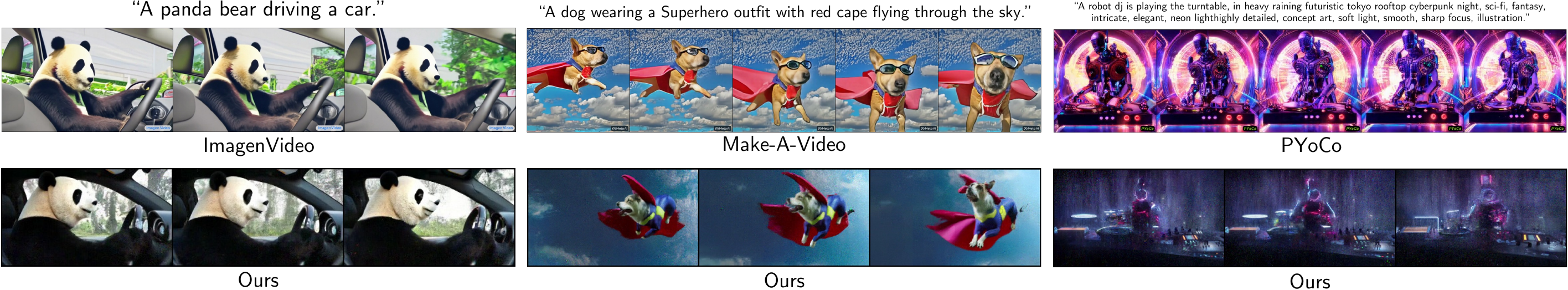}
\caption{\modelnameTTV is able to efficiently fine-tune from the standard low-resolution generator to high-resolution $64\times 288 \times 512$ text-to-video generation when fine-tuned from a low-resolution $36 \times 64$ diffusion for just 15,000 training steps.}
\label{fig:large-model-results}
\end{figure*}

%% file: tables/ucf-main.tex
\begin{table}[]
\caption{Comparison with the recent state-of-the-art methods on UCF-101~\cite{UCF101_dataset} $16 \times 256^2$ class-conditional video generation (note that our model is trained $64 \times 256^2$ videos). $^*$Note that Make-A-Video~\cite{Make-A-Video} was pretrained on a large-scale text-to-video dataset.}
\label{tab:ucf-main-results}
\centering
\begin{tabular}{lccc}
\toprule
Method & FVD $\downarrow$ & IS$\uparrow$ & Venue \\
\midrule
DIGAN~\cite{DIGAN} & 1630.2 & 29.71 & ICLR'22 \\
MoCoGAN-HD~\cite{MoCoGAN-HD} & 700 & 33.95 & ICLR'21 \\
StyleGAN-V~\cite{StyleGAN-V} & 1431.0 & 23.94 & CVPR'22 \\
TATS~\cite{Make-A-Video} & 635 & 57.63 & ECCV'22 \\
VIDM~\cite{VIDM} & 294.7 & - & AAAI'23 \\
PVDM~\cite{PVDM} & 343.6 & 74.4 & CVPR'23 \\
Make-A-Video$^*$~\cite{Make-A-Video} & \cellsecond{81.25} & 82.55 & ICLR'23 \\
\midrule
\modelnameS\ & 344.5 & 73.73 & \\
\modelnameM\ & 143.1 & \cellsecond{84.29} & CVPR'24 \\
\modelnameL\ & \cellbest{66.32} & \cellbest{87.68} & \\
\bottomrule
\end{tabular}
\end{table}

%% file: tables/ucf-ablations.tex
\begin{table}
\caption{Ablating architectural components in terms of FVD scores and training speed measured as the videos/sec throughpout on a single NVidia A100 80GB GPU.}
\label{tab:ucf-ablations}
\centering
\resizebox{1.0\linewidth}{!}{
\begin{tabular}{lcccc}
\toprule
\multirow{2}{*}{Setup} & \fvdmini & \fvdmini & \fvdmini & Training \\
 & ${16 \times 64^2}$ & ${32 \times 128^2}$ & ${64 \times 256^2}$ & speed $\uparrow$ \\
\midrule
Shallow fusion & 298.9 & 411.9 & 467.0 & 4.91 \\
Context detach & 290.6 & 375.0 & 397.3 & 4.4 \\
No adapt. computation & 319.3 & 391.5 & 373.9 & 2.73 \\
No coordinates & 305.3 & 400.7 & 389.5 & 4.47 \\
\midrule
Default model & 287.6 & 376.6 & 378.2 & 4.4 \\
\bottomrule
\end{tabular}
}
\end{table}

%% file: tables/ucf-inference-ablations.tex
\begin{table}[]
\caption{\fvdmini\ for various overlapped inference strategies.}
\label{tab:ucf-inference-ablations}
\centering
\begin{tabular}{lcc}
\toprule
Inference strategy & ${32 \times 128^2}$ & ${64 \times 256^2}$ \\
\midrule
No overlapping & 385.40 & 475.05 \\
50\% $w$-overlapping & 367.10 & 452.79 \\
50\% $h$-overlapping & 383.15 & 467.36 \\
50\% $h/w$-overlapping & 382.25 & 456.10 \\
50\% $f$-overlapping & 380.63 & 460.74 \\
50\% $f/w$-overlapping & 398.77 & 492.84 \\
50\% $f/h$-overlapping & 360.46 & 436.81 \\
50\% $f/h/w$-overlapping & 381.85 & 467.37 \\
\bottomrule
\end{tabular}
\end{table}

%% file: figures/overlapped-inference.tex
\begin{figure}
\includegraphics[width=\linewidth]{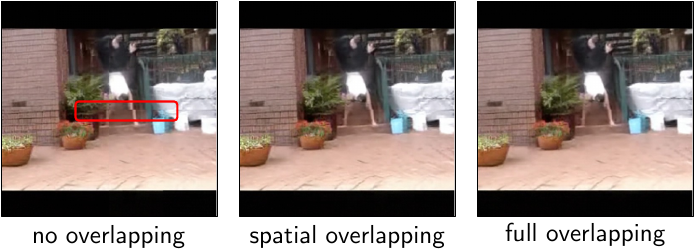}
\caption{Effect of the overlapped inference~\cite{MultiDiffusion}  on the consistency between the patches. Surprisingly, even without the full-resolution training~\cite{PatchDiffusion} and patch overlapping, our deep context fusion strategy manages to preserve strong consistency in the generated sample. See \cref{tab:ucf-inference-ablations} for quantitative analysis.}
\label{fig:overlapped-inference}
\end{figure}

%% file: tables/ucf-zero-shot.tex
\begin{table}[]
\caption{Zero-shot performance on UCF-101. \modelnameTTV\ achieves competitive performance when fine-tuned from the base low-resolution $36 \times 64$ generator for just 15,000 training steps.}
\label{tab:ucf-zero-shot}
\centering
\begin{tabular}{lccc}
\toprule
Method & Resolution & FVD$\downarrow$ & IS$\uparrow$ \\
\midrule
CogVideo~\cite{CogVideo} & $128 \times 128$ & 701.6 & 25.27 \\
Make-A-Video & $256\times256$ & 367.2 & 33.00 \\
MagicVideo~\cite{MagicVideo} & $256 \times 256$ & 655 & - \\
LVDM~\cite{LVDM} & $256 \times 256$ & 641.8 & - \\
Video LDM~\cite{VideoLDM} & N/A & 550.6 & 33.45 \\
VideoFactory~\cite{VideoFactory} & $256 \times 256$ & 410.0 & - \\
PYoCo~\cite{PYoCo} & $256 \times 256$ & 355.2 & 47.46 \\
\midrule
\modelnameTTV\ & $72 \times 128$ & 299.3 & 20.53 \\
\modelnameTTV\ & $144 \times 256$ & 383.3 & 21.15 \\
\modelnameTTV\ & $288 \times 512$ & 481.9 & 23.77 \\
\modelnameTTVK\ & $576 \times 1024$ & 447.5 & 24.51 \\
\bottomrule
\end{tabular}
\end{table}




%% file: sec/6_conclusion.tex
\section{Conclusion}
\label{sec:conclusion}

In this work, we developed the hierarchical patch diffusion model for high-resolution video synthesis, which efficiently trains in the end-to-end manner directly in the pixel space, and is amenable to swift fine-tuning from a base low-resolution diffusion model.
We showed state-of-the-art video generation performance on UCF-101, outperforming the recent methods by ${\approx}100\%$ in terms of FVD, and promising scalability results for text-to-video generation.
The techniques we developed hold significant potential for application across various patch-wise generative paradigms, including GANs, VAEs, autoregressive models, and beyond.
In future work, we intend to investigate better context conditioning, sampling strategies with stronger dependence enforcement, and also other tokenization/detokenization transformations to mitigate dead pixels artifacts.

%% file: supp/1_limitations.tex
\section{Limitations}
\label{supp:sec:limitations}

Although our model provides considerable improvements in video generation quality and enjoys a convenient end-to-end design, it still suffers from some limitations.

\inlinesection{Stitching artifacts}.
Despite using overlapped inference, our model occasionally exhibits stitching artifacts.
We illustrate these issues in \cref{fig:failure-cases} (left).
Inference strategies with stronger spatial communication, like classifier guidance~\cite{ADM}, should be employed to mitigate them.

\inlinesection{Error propagation}.
Since our model generally follows the cascaded pipeline~\cite{CDM, DALLE-2, GigaGAN, MDM} (with the difference that we train jointly and more efficiently), it suffers from the typical cascade drawback: the errors made in earlier stages of the pyramid are propagated to the next.
The error propagation artifacts are illustrated in \cref{fig:failure-cases} (left).

\inlinesection{Dead pixels}.
By ``dead pixels'' artifacts we imply failures of the ViT~\cite{ViT}-like pixel tokenization/detokenization procedure, where the model sometimes produces broken $4\times4$ patches.
They are illustrated in \cref{fig:failure-cases}.
These artifacts are unique to RINs~\cite{RIN} and we have not experienced them in our preliminary experiments with UNets~\cite{ADM, EDM}.
However, since they do not appear catastrophically often, we chose to continue to experiment with RINs.

\inlinesection{Slow inference}.
Patch-wise inference requires more function evaluations at test time, which slows down the inference process.
For our exponentially growing pyramid starting at $8 \times 36 \times 64$ and ending at $64 \times 288 \times 512$, with full (i.e., maximal) overlapping, we need to produce $(2 \cdot \frac{64}{8} - 1) \times (2 \cdot \frac{288}{36} - 1) \times (2 \frac{512}{64} - 1) = 3375$ patches for a single reverse diffusion step (see \cref{sec:method:sampling} for calculation details).
Adaptive computation with caching greatly accelerates this process, but it is still heavy.

\input{figures/failure-cases}

%% file: figures/failure-cases.tex
\begin{figure}[h]
\includegraphics[width=\linewidth]{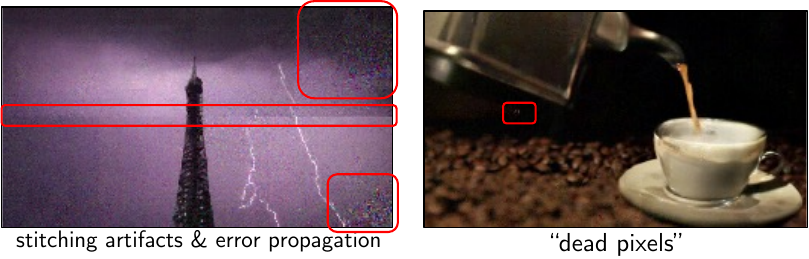}
\caption{Illustrating the failure cases of \modelname..}
\label{fig:failure-cases}
\end{figure}

%% file: supp/2_additional_results.tex
\section{Additional results}
\label{supp:sec:additional-results}

There are multiple incosistencies in quantitative evaluation of video generators that are inconsistent between previous projects~\cite{StyleGAN-V, PVDM}.
For FVD~\cite{FVD} on UCF101 (the most popular metric for it), there are differences in the amounts of fake/real videos used to compute the statistics, FPS values, resolutions, and real data subsets (``train'' or ``train + test'').
To account for these differences, in \cref{tab:additional-results}, we release a comprehensive set of metrics for easier assessment of our models' performance in comparison with the prior work.
Apart from that, it also includes additional models, HPDM-S and HPDM-M, and also the results for the fixed version of our text-to-video HPDM model (after the main deadline, we noticed that our FSDP-based~\cite{FSDP} training was not updating some of the EMA parameters properly, which was the cause of gaussian jitter artifacts in \cref{fig:failure-cases}).



To compute real data FVD statistics, we always use the train set of UCF-101 (around 9.5k videos in total).
We train the models with the default 25FPS resolution.
Our models are trained for 64 frames, and to compute the results for $16$ frames, we simply take the first 16 frames out of the sequence.

\input{tables/additional-results}
\input{tables/additional-results-zero-shot}
\input{figures/fvd-correlations}

\input{figures/large-ucf-vis}
\input{figures/t2v-many}

Additional results are also provided on the project webpage: \projecthref.

%% file: tables/additional-results.tex
\begin{table*}[h]
\caption{Additional FVD evaluation results for class-conditional UCF-101 video generation. ``Pre-trained'' denotes whether the model was pre-trained on an external dataset. ``\#samples'' is the amount of fake videos used to compute the fake data statistics. In \cref{fig:fvd-correlations}, we also demonstrated that FVD scores computed for different amount of samples are well-correlated with one another. For IS, we cannot compute it for 64-frames-long videos due to the design of C3D model~\cite{TGANv2, StyleGAN-V}.}
\label{tab:additional-results}
\centering
\begin{tabular}{lcccccc}
\toprule
Method & Resolution & Pre-trained? & \#samples & FVD$\downarrow$ & IS$\uparrow$ \\
\midrule
DIGAN~\cite{DIGAN} & $16 \times 128 \times 128$ & \xmark & 2,048 & 1630.2 & 00.00 \\
StyleGAN-V~\cite{StyleGAN-V} & $16 \times 256 \times 256$ & \xmark & 2,048 & 1431.0 & 23.94 \\
TATS~\cite{TATS} & $16 \times 128 \times 128$ & \xmark & N/A & 332 & 79.28 \\
VIDM~\cite{VIDM} & $16 \times 256 \times 256$ & \xmark & 2,048 & 294.7 & - \\
LVDM~\cite{LVDM} & $16 \times 256 \times 256$ & \xmark & 2,048 & 372 & - \\
PVDM~\cite{PVDM} & $16 \times 256 \times 256$ & \xmark & 2,048 & 343.6 & - \\
PVDM~\cite{PVDM} & $16 \times 256 \times 256$ & \xmark & 10,000 & - & 74.40 \\
PVDM~\cite{PVDM} & $128 \times 256 \times 256$ & \xmark & 2,048 & 648.4 & - \\
VideoFusion~\cite{VideoFusion} & $16 \times 128 \times 128$ & \xmark & N/A & 173 & 80.03 \\
Make-A-Video$^*$~\cite{Make-A-Video} & $16 \times 256 \times 256$ & \cmark & 10,000 & \cellsecond{81.25} & 82.55 \\
\midrule
\multirow{4}{*}{\modelnameS} & $16 \times 256 \times 256$ & \xmark & 2,048 & 370.50 & 61.50 \\
& $16 \times 256 \times 256$ & \xmark & 10,000 & 344.54 & 73.73 \\
& $64 \times 256 \times 256$ & \xmark & 2,048 & 647.48 & N/A \\
& $64 \times 256 \times 256$ & \xmark & 10,000 & 578.80 & N/A \\
\midrule
\multirow{4}{*}{\modelnameM} & $16 \times 256 \times 256$ & \xmark & 2,048 & 178.15 & 69.76 \\
& $16 \times 256 \times 256$ & \xmark & 10,000 & 143.06 & \cellsecond{84.29} \\
& $64 \times 256 \times 256$ & \xmark & 2,048 & 324.72 & N/A \\
& $64 \times 256 \times 256$ & \xmark & 10,000 & 257.65 & N/A \\
\midrule
\multirow{4}{*}{\modelnameL} & $16 \times 256 \times 256$ & \xmark & 2,048 & 92.00 & 71.16 \\
& $16 \times 256 \times 256$ & \xmark & 10,000 & \cellbest{66.32} & \cellbest{87.68} \\
& $64 \times 256 \times 256$ & \xmark & 2,048 & 137.52 & N/A \\
& $64 \times 256 \times 256$ & \xmark & 10,000 & 101.42 & N/A \\
\bottomrule
\end{tabular}
\end{table*}

%% file: tables/additional-results-zero-shot.tex
\begin{table}[h]
\caption{Additional zero-shot FVD evaluation results for UCF-101. For zero-shot evaluation, to the best of our knowledge, all the prior works use 10,000 generated videos to compute the I3D statistics.}
\label{tab:additional-results-zeroshot}
\centering
\begin{tabular}{lccc}
\toprule
Method & Resolution & FVD$\downarrow$ & IS$\uparrow$ \\
\midrule
CogVideo~\cite{CogVideo} & $16 \times 480 \times 480$ & 701.6 & 25.27 \\
Make-A-Video & $16 \times 256\times256$ & 367.2 & 33.00 \\
MagicVideo~\cite{MagicVideo} & $16 \times 256 \times 256$ & 655 & - \\
LVDM~\cite{LVDM} & $16 \times 256 \times 256$ & 641.8 & - \\
Video LDM~\cite{VideoLDM} & N/A & 550.6 & 33.45 \\
VideoFactory~\cite{VideoFactory} & $16 \times 256 \times 256$ & 410.0 & - \\
PYoCo~\cite{PYoCo} & $16 \times 256 \times 256$ & 355.2 & 47.46 \\
\midrule
\multirow{5}{*}{HPDM-T2V}
& $16 \times 144 \times 256$ & 383.26 & 21.15 \\
& $16 \times 256 \times 256$ & 728.26 & 23.46 \\
& $16 \times 288 \times 512$ & 481.93 & 23.77 \\
& $64 \times 256 \times 256$ & 1238.62 & N/A \\
& $64 \times 288 \times 512$ & 1197.60 & N/A \\
\bottomrule
\end{tabular}
\end{table}

%% file: figures/fvd-correlations.tex
\begin{figure}
\centering
\includegraphics[width=0.6\linewidth]{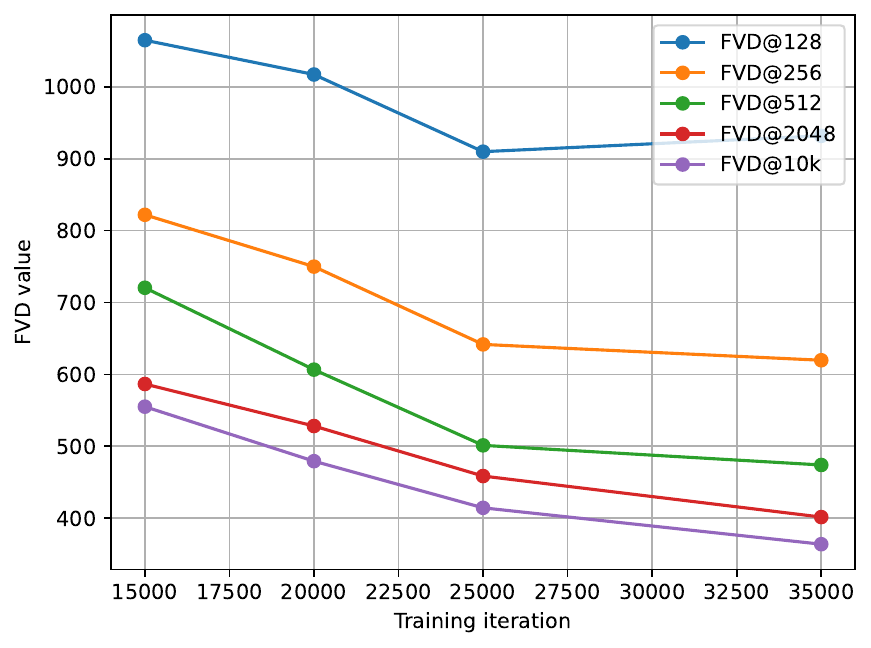}
\caption{Using different amounts of fake videos to compute FVD~\cite{FVD} gives very correlated, but offset values with the main trend being ``the more ---- the better''. We hypothesize that using more synthetic samples yields better coverage of different modes of the data distribution and decreases the influence of outliers. These FVD scores are computed for different training steps of \modelnameS. Using too few videos leads to undiscriminative results only closer to convergence.}
\label{fig:fvd-correlations}
\end{figure}

%% file: figures/large-ucf-vis.tex
\begin{figure*}
\centering
\includegraphics[width=0.9\linewidth]{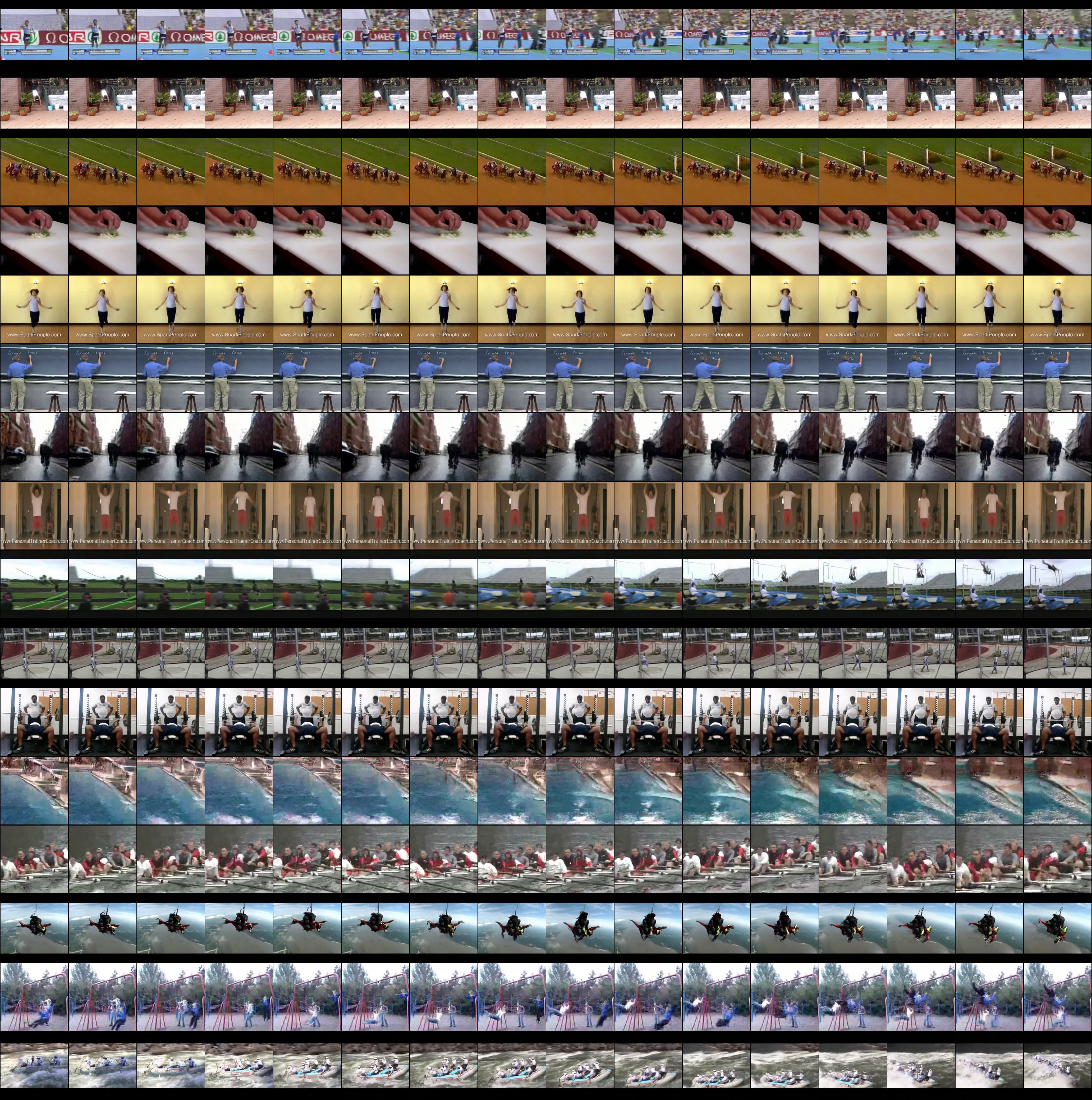}
\caption{\emph{Random} samples from \modelnameL\ on UCF-101 $64\times256^2$~\cite{UCF101_dataset} without classifier-free guidance. We display 16 frames from a 64-frame-long video with $4\times$ subsampling.}
\label{fig:large-ucf-vis}
\end{figure*}

%% file: figures/t2v-many.tex
\begin{figure*}
\centering
\centering
\begin{subfigure}[b]{0.9\textwidth}
    \centering
    \includegraphics[width=\textwidth]{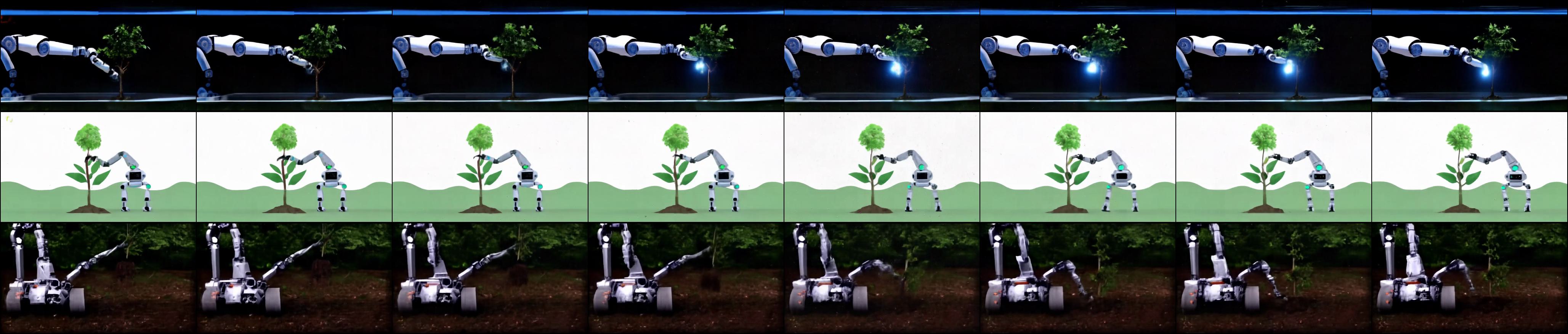}
    \caption{``A robot planting a tree.''}
\end{subfigure}
\hfill
\begin{subfigure}[b]{0.9\textwidth}
    \centering
    \includegraphics[width=\textwidth]{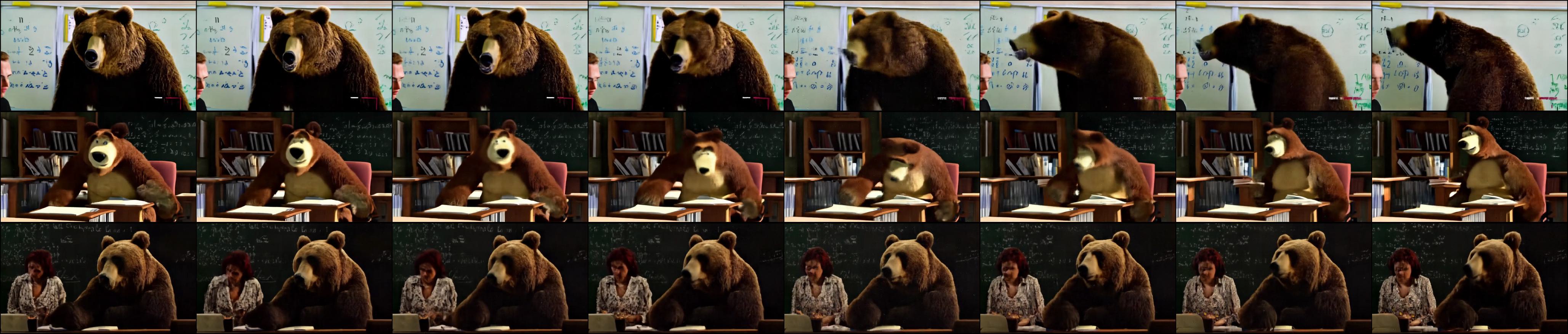}
    \caption{``A confused grizzly bear in calculus class.''}
\end{subfigure}
\hfill
\begin{subfigure}[b]{0.9\textwidth}
    \centering
    \includegraphics[width=\textwidth]{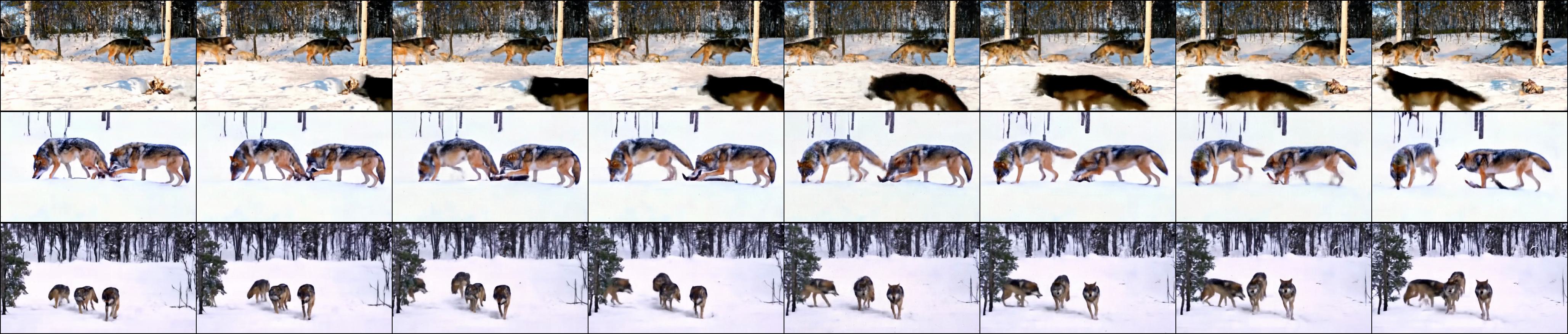}
    \caption{``A high-definition video of a pack of wolves hunting in a snowy forest, natural behavior, dynamic angles.''}
\end{subfigure}
\hfill
\begin{subfigure}[b]{0.9\textwidth}
    \centering
    \includegraphics[width=\textwidth]{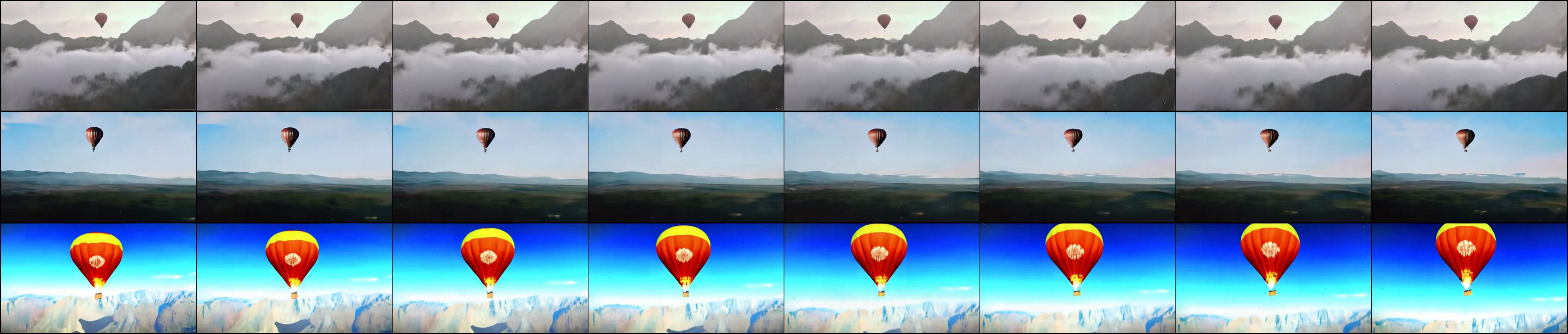}
    \caption{``A hot air balloon floating over a mountain range.''}
\end{subfigure}
\caption{Text-to-video generation results for variable text prompts. Note that our text-to-video model has been fine-tuned only for 15k training steps from a $36\times 64$ low-resolution generator. Animations and comparisons to the current SotA can be found in the supplementary.}
\label{fig:t2v-many}
\end{figure*}

%% file: supp/3_impl_details.tex
\section{Implementation details}
\label{sec:impl-details}

In this section, we provide additional implementation details for our model.
We train our model in a patch-wise fashion with the patch resolution of $16\times 64\times 64$ for UCF-101~\cite{UCF101_dataset} and $8\times 36\times 64$ for text-to-video generation.
After the main deadline, we continued training our model on UCF for several more training steps, and also trained two smaller versions for fewer steps.
We denote the smaller versions as \modelnameS\ and \modelnameM, while the larger one is denoted as \modelnameL.
They differ in the amount of training steps performed and also the latent dimensionality of RINs~\cite{RIN}: 256, 512 and 1024, respectively.
Our text-to-video model \modelnameTTV\ was fine-tuned for 15k steps and \modelnameTTVK\ for 100k steps.
We provide the hyperparameters for our models in \cref{tab:hyperparameters}.
For sampling, we use spatial 50\% patch overlapping to compute the metrics (for performance purposes), and full overlapping for visualizations.
We use stochastic sampling with second-order correction~\cite{EDM} for the first pyramid level.
For later stages, we use 
Also, we disabled stochasticity for text-to-video synthesis since we have not observed it to be improving the results.
We use 128 steps for the first pyramid stage, and then decrease them exponentially for later stages, dividing the number of steps by 2 with each pyramid level increase.

\input{figures/architecture-full}
\input{tables/hyperparameters}

%% file: figures/architecture-full.tex
\begin{figure*}
\centering
\includegraphics[width=0.7\linewidth]{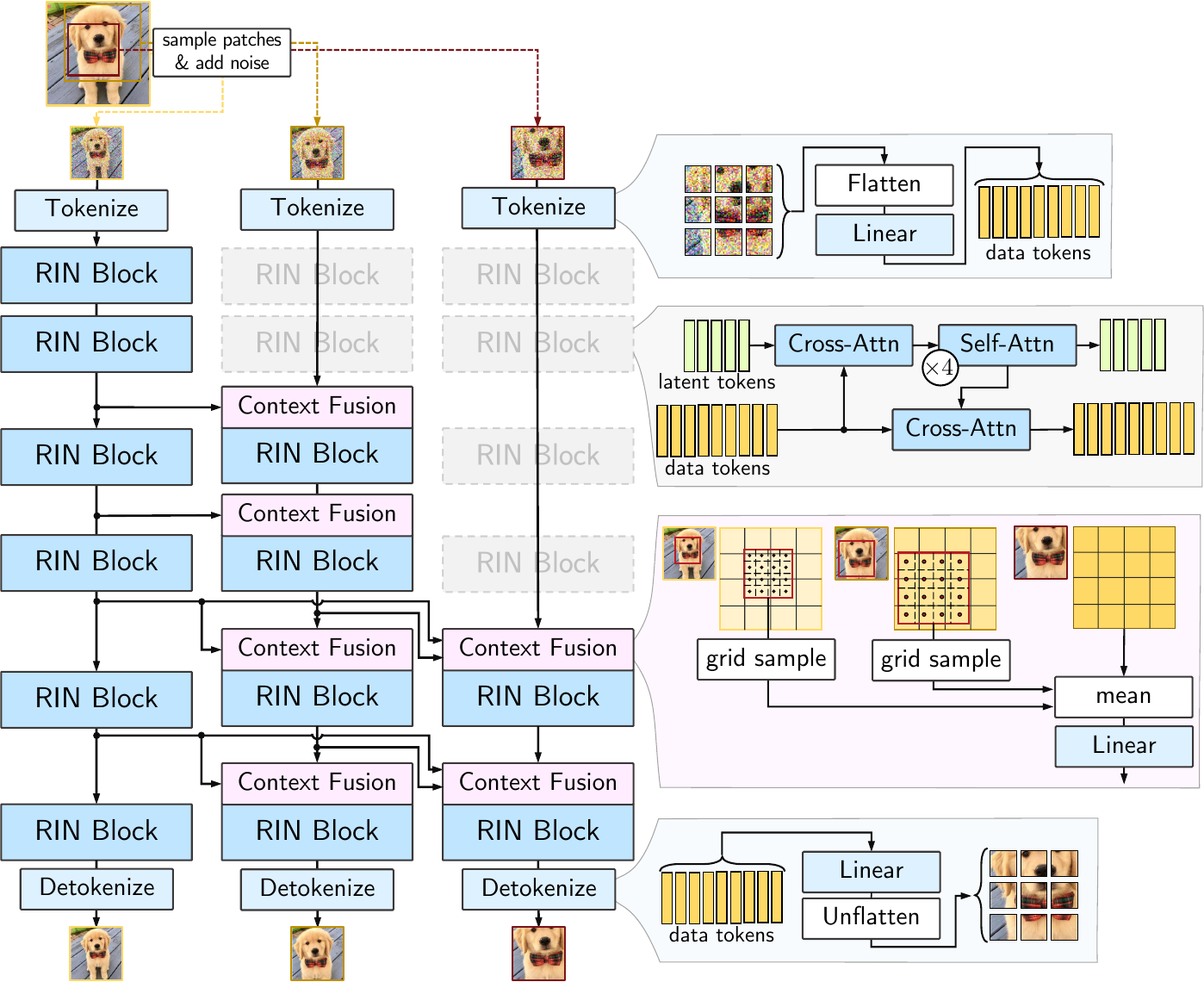}
\caption{Full architecture illustration of \modelname with depiction of the blocks.}
\label{fig:architecture-full}
\end{figure*}

%% file: tables/hyperparameters.tex
\begin{table*}
\caption{Hyperparameters for different variations of \modelname. For all the models, we used almost the same amount hyperparameters. For \modelnameTTV, we used joint video + image training which is reflected by its batch size. For \modelnameTTV and \modelnameTTVK, we also used low-res pre-training by first training the lowest pyramid stage on $36 \times 64$-resolution videos for 500k steps.}
\label{tab:hyperparameters}
\centering
\resizebox{1.0\linewidth}{!}{
\begin{tabular}{lccccc}
\toprule
Hyperparameter & \modelnameS & \modelnameM & \modelnameL & \modelnameTTV & \modelnameTTVK \\
\midrule
Conditioning information & class labels & class labels & class labels & T5-11B embeddings & T5-11B embeddings \\
Conditioning dropout probability & 0.1 & 0.1 & 0.1 & 0.1 & 0.1 \\
Tokenization dim & 1024 & 1024 & 1024 & 1024 & 1024 \\
Tokenizer resolution & $1\times 4 \times 4$ & $1\times 4 \times 4$ & $1\times 4 \times 4$ & $1\times 3 \times 4$ & $1\times 3 \times 4$ \\
Latent dim & 256 & 512 & 1024 & 3072 & 3072 \\
Number of latents & 768 & 768 & 768 & 768 & 768 \\
Batch size & 768 & 768 & 768 & 4096 + 4096 & 1024 + 1024 \\
Target LR & 0.005 & 0.005 & 0.005 & 0.005 & 0.005 \\
Weight decay & 0.01 & 0.01 & 0.01 & 0.01 & 0.01 \\
Number of warm-up steps & 10k & 10k & 10k & 5k & 5k \\
Parallelization strategy & DDP & DDP & DDP & FSDP & FSDP \\
Starting resolution & $16\times 64 \times 64$ & $16\times 64 \times 64$ & $16\times 64 \times 64$ & $8\times 36 \times 64$ & $16\times 72 \times 128$ \\
Target resolution & $64\times 256 \times 256$ & $64\times 256 \times 256$ & $64\times 256 \times 256$ & $64\times 288 \times 512$ & $16\times 576 \times 1024$ \\
Patch resolution & $16\times 64 \times 64$ & $16\times 64 \times 64$ & $16\times 64 \times 64$ & $8\times 36 \times 64$ & $16\times 72 \times 128$ \\
Number of RIN blocks~\cite{RIN} & 6 & 6 & 6 & 6 & 6 \\
Number of pyramid levels & 3 & 3 & 3 & 4 & 4 \\
Number of pyramid levels per block & 1/1/2/2/3/3 & 1/1/2/2/3/3 & 1/1/2/2/3/3 & 1/2/2/3/3/4 & 4/4/4/4/4/4 \\
\midrule
Number of parameters & 178M & 321M & 725M & 3,934M & 3,934M \\
Number of training steps & 40k & 40k & 65k & 15k (+ 500k) & 100k (+ 500k) \\
\bottomrule
\end{tabular}
}
\end{table*}

%% file: supp/4_failed_experiments.tex
\section{Failed experiments}
\label{supp:sec:failed-experiments}

In this section, we provide a list of ideas which looked promising inutitively, but didn't work out at the end --- either because of some fundamental fallacies related to them, or the lack of experimentation and limited amount of time to explore them, or because of some potential implementation bugs which we have not been aware of.

\begin{enumerate}
    \item \emph{Cached inference has not sped up inference as much as we expected}. As described in \cref{sec:method:misc} and \cref{sec:impl-details}, we cache the activations from previous pyramid levels when sampling its higher stages. However, the speed-up was just ${\approx}40\%$, which was not decisive. One issue is that we do not cache some activations (tokenizer activations and contexts). But the other reason is that grid-sampling is expensive. Grid sampling could be avoided by upsampling and then slicing, but this would lead to additional memory usage and will complicate the inference code.
    \item \emph{Positional encoding of the coordinates}. For some reason, the model started to diverge when we tried replacing raw coordinates with their sinusodial embeddings. We believe that this direction is still promising, but is under-explored.
    \item \emph{Stochastic sampling and second-order sampling for later stages}. For UCF-101, we use stochastic sampling for the first pyramid level, but disabled it for text-to-video generation. Also, second-order correction was producing grainy artifacts for later pyramid stages.
    \item \emph{Weight sharing between blocks}. To conserve GPU memory, we tried to share the weights between all the transformer blocks, but that led to inferior results.
    \item \emph{Cheap high-res + expensive low-res U-Net backbone}. U-Nets were also not converging well for us in their regular design and were not giving substantial performance yields when combined with adaptive computation (only ${\approx}$10\% during training versus ${\approx}$50\% in RINs) due to the irregular amounts of blocks per resolution in their design.
    \item \emph{Random pyramid cuts}. Another strategy to make the later pyramid stages cheaper during training was to compute them only once in a while. For this, we would randomly sample the amount of pyramid stages for each mini batch per GPU. When parallelizing across many GPUs, this strategy gives enough randomness. While it decreased the training costs without severe quality degradation, it does not speed up inference and complicates logging.
    \item \emph{Mixed precision training}. It produced consistently worse convergence, either with manual mixed precision or autocast, either for FP16 and BF16.
    \item \emph{Fusing patch features for all the layers}. That strategy was not giving much quality improvement, but was tremendously expensive, which is why we gave it up.
\end{enumerate}

%% file: supp/5_societal-impact.tex
\section{Potential negative impact}
\label{sec:societal-impact}

We introduced a patch-wise diffusion-based video generation model: a new paradigm for video generation that is a step forward in the field.
While our model exhibits promising capabilities, it's essential to consider its potential negative societal impacts:



\begin{itemize}
    \item \emph{Misinformation and Deepfakes}. While our text-to-video model underperforms compared to the largest existing ones (.e.g, \cite{ImagenVideo, Make-A-Video}), it demonstrates a promising direction on how to improve the existing generators further, which creates a risk of generative AI misuse in creating misleading videos or deepfakes. This can contribute to the spread of misinformation or be used for malicious purposes.
    \item \emph{Intellectual Property Concerns}. The ability to generate videos can lead to challenges in copyright and intellectual property rights, especially if the technology is used to replicate or modify existing copyrighted content without permission.
    \item \emph{Economic Impact}. Automation of video content generation could impact jobs in industries reliant on manual content creation, leading to economic shifts and potential job displacement.
    \item \emph{Bias and Representation}. Like any AI model, ours is subject to the biases present in its training data. This can lead to issues in representation and fairness, especially if the model is used in contexts where diversity and accurate representation are crucial.
\end{itemize}

To address the potential negative impacts, it is crucial to:
\begin{itemize}
    \item Develop and enforce strict ethical guidelines for the use of video generation technology.
    \item Continuously work on improving the model to reduce biases and ensure fair representation.
    \item Collaborate with legal and ethical experts to understand and navigate the implications of video synthesis technology in terms of intellectual property rights.
Engage with stakeholders from various sectors to assess and mitigate any economic impacts, particularly concerning job displacement.

\end{itemize}
In conclusion, while our model represents a notable advancement in video generation technology, it is imperative to approach its deployment and application with a balanced perspective, considering both its benefits and potential societal implications.